\pdfoutput=1
\documentclass[10pt,letterpaper]{article}
\usepackage{cogsci}
\usepackage{pslatex}
\usepackage{apacite}
\usepackage{comment}
\usepackage{graphicx}
\usepackage{graphbox}
\usepackage{nicefrac}
\usepackage{lipsum}
\usepackage{url}
\usepackage[T1]{fontenc}

\usepackage{amsmath,amssymb}
\DeclareMathOperator*{\argmax}{arg\,max}

\title{Where Do Heuristics Come From?}
 
\author{{\large \bf Marcel Binz (binz@staff.uni-marburg.de)} \\
  Department of Psychology, Theoretical Neuroscience Group \\
  Philipps-Universit\"at Marburg
  \AND {\large \bf Dominik Endres (dominik.endres@uni-marburg.de)} \\
  Department of Psychology, Theoretical Neuroscience Group \\
  Philipps-Universit\"at Marburg}

\begin{document}

\maketitle

\begin{abstract}
Human decision-making deviates from the optimal solution, i.e. the one maximizing cumulative rewards, in many situations. Here we approach this discrepancy from the perspective of computational rationality and our goal is to provide justification for such seemingly sub-optimal strategies. More specifically we investigate the hypothesis, that humans do not know optimal decision-making algorithms in advance, but instead employ a learned, resource-constrained approximation. The idea is formalized through combining a recently proposed meta-learning model based on Recurrent Neural Networks with a resource-rational objective. The resulting approach is closely connected to variational inference and the Minimum Description Length principle. Empirical evidence is obtained from a two-armed bandit task. Here we observe patterns in our family of models that resemble differences between individual human participants. 

\end{abstract}

\section{Introduction}

In this work we study human decision-making strategies on a stationary multi-armed bandit task. These are among the simplest sequential decision-making problems, that require reasoning about trade-offs between exploration and exploitation. In the special case of an infinite horizon and geometric discounting their Bayes-optimal solution is the Gittins index strategy \cite{gittins1979bandit}, while in general it is defined as the result of a planning process in an augmented Markov Decision Process \cite{duff2002optimal}. Prior work however suggests, that several heuristics appear to be favourable as a model of human decision-making, when compared to the Bayes-optimal solution \cite{steyvers2009bayesian, zhang2013forgetful}.  \\

Understanding human cognition in terms of heuristics has been a major theme in cognitive science over the past decades \cite{tversky1974judgment, simon1990invariants, gigerenzer1999simple}. They can be viewed as crude, but realizable, approximations of optimal behavior. Heuristics are thus connected to the idea of rationality under resource constraints, which is commonly referred to as bounded rationality \cite{simon1972theories}, computational rationality \cite{gershman2015computational}, or resource-rationality \cite{griffiths2015rational}. Examples for resource constraints include related prior experience on a given task, limited capacity of our brain or restricted deliberation times. For a more general overview of computational rationality we refer the reader to \citeA{gershman2015computational}. Here we are interested in the hypothesis, that humans employ a learned, resourced-constrained approximation of an optimal decision-making strategy. More specifically we show, that different, potentially sub-optimal, human strategies emerge naturally in artificial learning systems when varying the strength of the constraints placed upon them. For a realization of this principle, we rely on information-theoretic concepts, similar to the approach of \citeA{ortega2013thermodynamics}.  \\

We instantiate a particular kind of such resource-rational agents using recent advances from the meta-learning literature \cite{wang2016learning, duan2016rl}. In this framework the algorithm to be learned is parametrized by a Recurrent Neural Network (RNN). RNNs are known to be Turing-complete and hence are in theory able to realize any algorithm \cite{siegelmann1991turing}. The RNN is trained on a set of related tasks to act as an independent Reinforcement Learning algorithm for solving the original problem. We treat all parameters of the RNN as random variables and infer approximate posterior distributions by solving a regularized optimization problem. Varying the regularization factor leads to a spectrum of resource-rational algorithms, each possessing different properties. Models with large constraints need to rely more on prior assumptions and thus prefer simple strategies, while models with weaker constraints will approach the optimal solution (up to the representational capabilities of the RNN and the limitations of the meta-learning procedure).\\

The resulting approach is closely related to the Minimum Description Length (MDL) principle \cite{hinton1993keeping, grunwald2004tutorial}, which asserts that the best model is the one, that leads to the best compression of the data, including a cost for describing the model. The bits-back argument establishes a link between the MDL principle and Bayesian learning \cite{honkela2004variational}, opening up connections to Bayesian theories of cognition \cite{griffiths2008bayesian}. Indeed several heuristics have been recently interpreted as Bayesian models under strong priors \cite{parpart2018heuristics}. \\

Our hypothesis is validated on a classical two-armed bandit task. However we view multi-armed bandits merely as the first step towards investigating more complex tasks and the proposed algorithm is not limited to any specific problem class. The following section first introduces the framework in more general terms, before considering multi-armed bandits as a special case. We then identify different strategies of human participants and subsequently show how the proposed class of models captures important characteristics of human behavior on both a qualitative and quantitative level. Our results indicate, that the seemingly sub-optimal decision strategies used by humans might be a consequence of the constraints under which these very strategies are learned.

\section{Methods}

\subsection{Reinforcement Learning}

Let $M = (\mathcal{S}, \mathcal{A}, p, \gamma)$ be a Markov Decision Process (MDP), with a set of states $\mathcal{S}$, a set of actions $\mathcal{A}$, a joint distribution over the next state and a scalar reward signal, describing the dynamics of the environment, $p(s_{t+1}, r_t | s_t, a_t)$ and a discount factor $\gamma \in [0, 1]$. The objective of a Reinforcement Learning (RL) agent is to find a policy $\pi(a_t | \cdot)$, that maximizes the discounted, expected return $\mathbb{E}_{p, \pi} \left[ \sum_{t=0}^{\infty} \gamma^t r_t \right]$ without having direct access to the true underlying dynamics $p$. 

\subsection{Learning Reinforcement Learning Algorithms}

Following the approach of \citeA{wang2016learning, duan2016rl} we want to \emph{learn} a RL algorithm for solving a MDP sampled from a distribution over MDPs. We parametrize the algorithm to be learned with a Recurrent Neural Network (RNN), in form of a Gated Recurrent Unit \cite{cho2014learning}, followed by a linear layer. The set of all model parameters is denoted with $\theta$ in the following. The RNN takes previous actions and rewards as inputs in addition to the current state, making the output a function of the entire history $X_t = (s_0, a_0, r_0, s_1 \ldots, a_{t-1}, r_{t-1}, s_t)$. A good algorithm has to integrate information from the history in order to identify the currently active MDP, based on which it subsequently has to select the appropriate strategy. The RNN is trained to accomplish this using standard model-free RL techniques. In this work we utilize $n$-step Q-Learning \cite{mnih2016asynchronous}, although in theory any other algorithm could be applied as well. The RNN implements a freestanding RL algorithm through its recurrent activations after training is completed (the parameters of the RNN are held constant during evaluation). Throughout this work we use the abbreviation LRLA -- for learned Reinforcement Learning algorithm -- to refer to this kind of model. Alternatively we can view this procedure as a model-free algorithm for partially observable MDPs, where the hidden information consists of the currently active task.

\subsection{Resource-Rational Decision-Making} 

We consider maximizing the following regularized objective for inferring a distribution $q_{\phi}$ over parameters $\theta$ of LRLAs: 
\begin{align} \label{eq:constraint}
    \mathcal{L}(\phi, \mathbf{X}, \mathbf{y}) = \mathbb{E}_{q_{\phi}(\theta)} \left[ \log p(\mathbf{y} | \mathbf{X}, \theta) \right] - \beta \text{KL}(q_{\phi}(\theta) || p(\theta))
\end{align}

where the hyperparameter $\beta$ controls how much the posterior is allowed to deviate from the prior in terms of the Kullback-Leibler (KL) divergence. We assume a likelihood $p(\mathbf{y} | \mathbf{X}, \theta)$, that factorizes over data points $\prod_{i=1}^{N} p(y_i | X_i, \theta)$ and we approximate each factor with a normal distribution of fixed scale $\sigma_y$: $\mathcal{N}(y_t; {Q}_{\theta}(X_t, a), \sigma_{y})$. In our setting ${Q}_{\theta}(X_t, a)$ corresponds to the RNN output after seeing history $X_t$ and $y_t$ corresponds to the n-step return  $\sum_{k=0}^{n-1} \gamma^k r_{t+k} + \gamma^n \max_{a}Q_{\theta}(X_{t+n}, a)$. The corresponding policy is derived as follows:
\begin{equation}\label{eq:pol}
    \pi(a_t | X_t) = \begin{cases} 1 & \text{if } a_t = \argmax_{a \in \mathcal{A}} Q_{\theta}(X_{t}, a) \\ 0 & \text{else} \end{cases}
\end{equation}

Setting $\beta$ to a specific value can be interpreted as implicitly defining a constraint on $\text{KL}(q_{\phi}(\theta) || p(\theta))$. Importantly the KL term determines how much the model parameters can be compressed in theory \cite{hinton1993keeping}. Hence our models are resource-constrained with regard to a hypothetical lower bound on their storage capacity. Intuitively, if the regularization factor $\beta$ is large, parameters are forced to match the prior closely. In this work we employ priors favoring simple functions, hence models are only allowed to realize more complex functions as $\beta \rightarrow 0$. \\

\subsection{Bayesian Interpretation}

If we set $\beta = 1$, we recover the evidence lower bound (ELBO) as an objective for performing variational inference. In the setting of large data-sets subsampling techniques are often employed to approximate Equation \ref{eq:constraint} using mini-batches $\mathcal{B}$ of size M with an appropriately scaled log-likelihood term:
\begin{equation} \label{eq:N}
    \log p(\mathbf{y} | \mathbf{X}, \theta) \approx \frac{N}{M}  \sum_{i \in \mathcal{B}} \log p(y_i | X_i, \theta)
\end{equation}

If data arrives in sequential fashion, as it does in the RL setting, the data-set size $N$ is not known in advance and has to be treated as an additional hyperparameter. This leads to a Bayesian interpretation of Equation \ref{eq:constraint} even for $\beta \neq 1$. For any values of $\beta$ and $N$ maximizing Equation \ref{eq:constraint} is equivalent to performing stochastic variational inference with an assumed data-set size of $\hat{N} = \frac{N}{\beta}$. In practice we optimize a by $N^{-1}$ scaled version of Equation \ref{eq:constraint}, which leads to $\hat{N}^{-1}$ as a factor for the KL term. \\

In the following section we investigate whether we can understand individual differences in human decision-making in terms of optimal solutions to Equation \ref{eq:constraint} for varying values of $\beta$. It is worth clarifying, that we are only interested in the computational aspects of this hypothesis, i.e. we want to test, whether human decision-making can be characterized through resource-rational strategies. We do not attempt to answer how this objective is realized on an algorithmic or implementational level.

\subsection{Technical Details}

\begin{figure*}[t]
    \centering
    \begin{tabular}{cccc}
      \parbox{3.5cm}{\centering \textbf{Value-directed} \\ \vspace*{0.5cm} $\pi(a_t = 0) = \Phi (V_t) $} &  \parbox{3.5cm}{\centering \textbf{Thompson Sampling} \\ \vspace*{0.5cm} $\pi(a_t = 0) = \Phi ({V}_t/{TU}_t) $} & \parbox{3.5cm}{\centering \textbf{UCB} \\ \vspace*{0.5cm} $\pi(a_t = 0) = \Phi (V_t + RU_t) $} & \parbox{3.5cm}{\centering \textbf{LRLA ($\beta = 0$)} \\ \vspace*{0.5cm} Equation \ref{eq:pol}} \\ 
      & & & \\
      \includegraphics[align=c, width=0.23\textwidth]{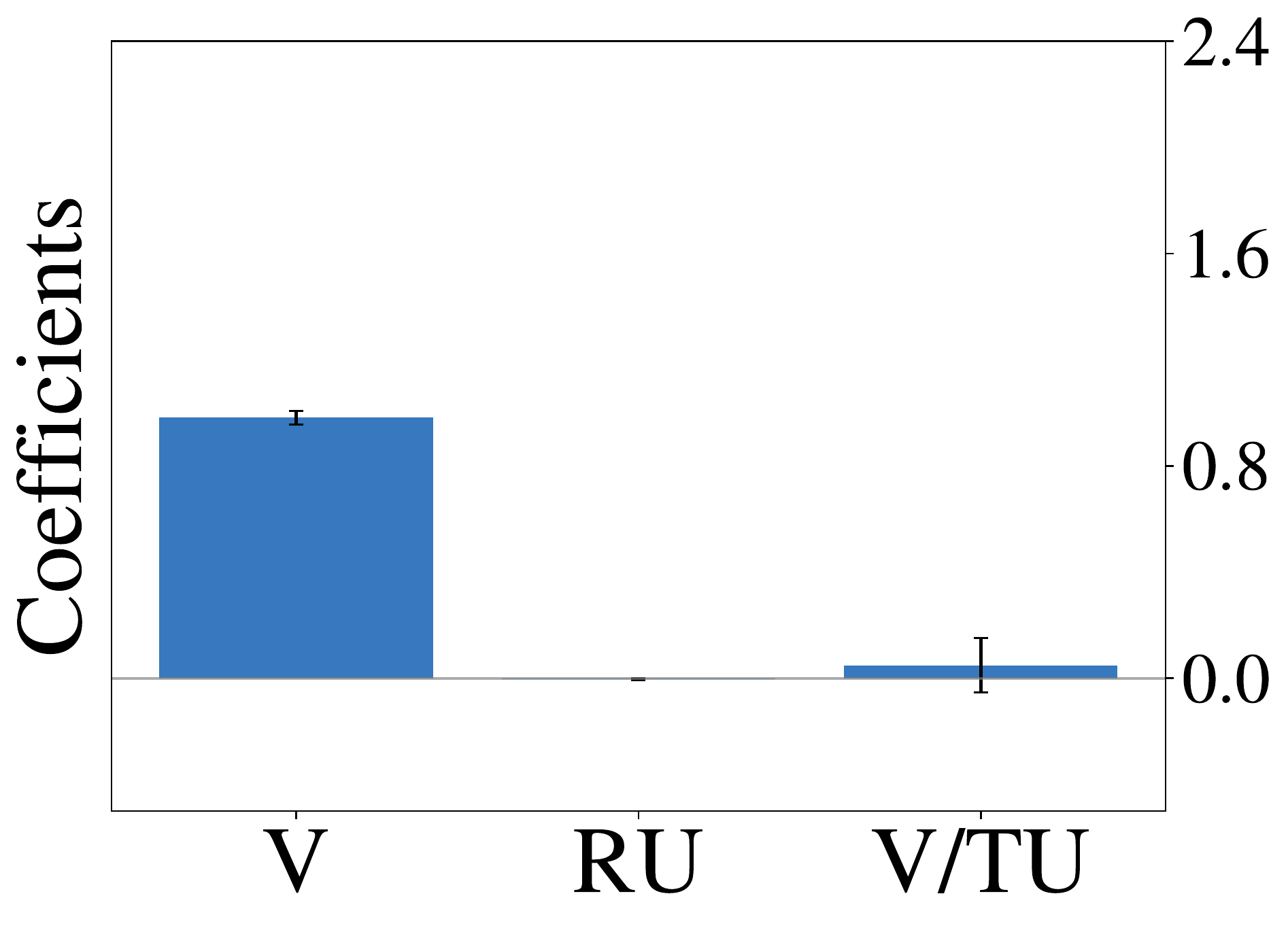} & \includegraphics[align=c, width=0.23\textwidth]{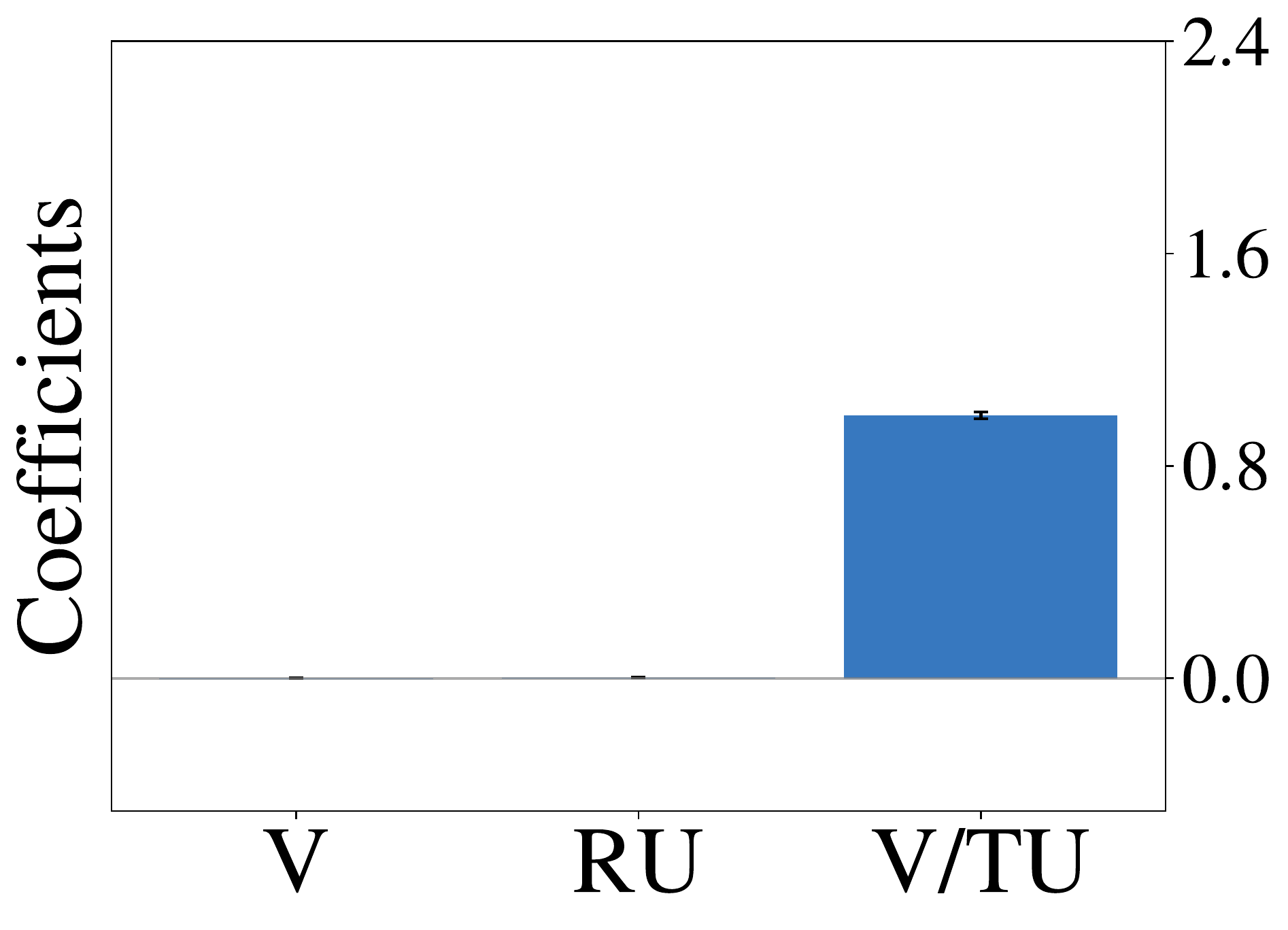} & \includegraphics[align=c, width=0.23\textwidth]{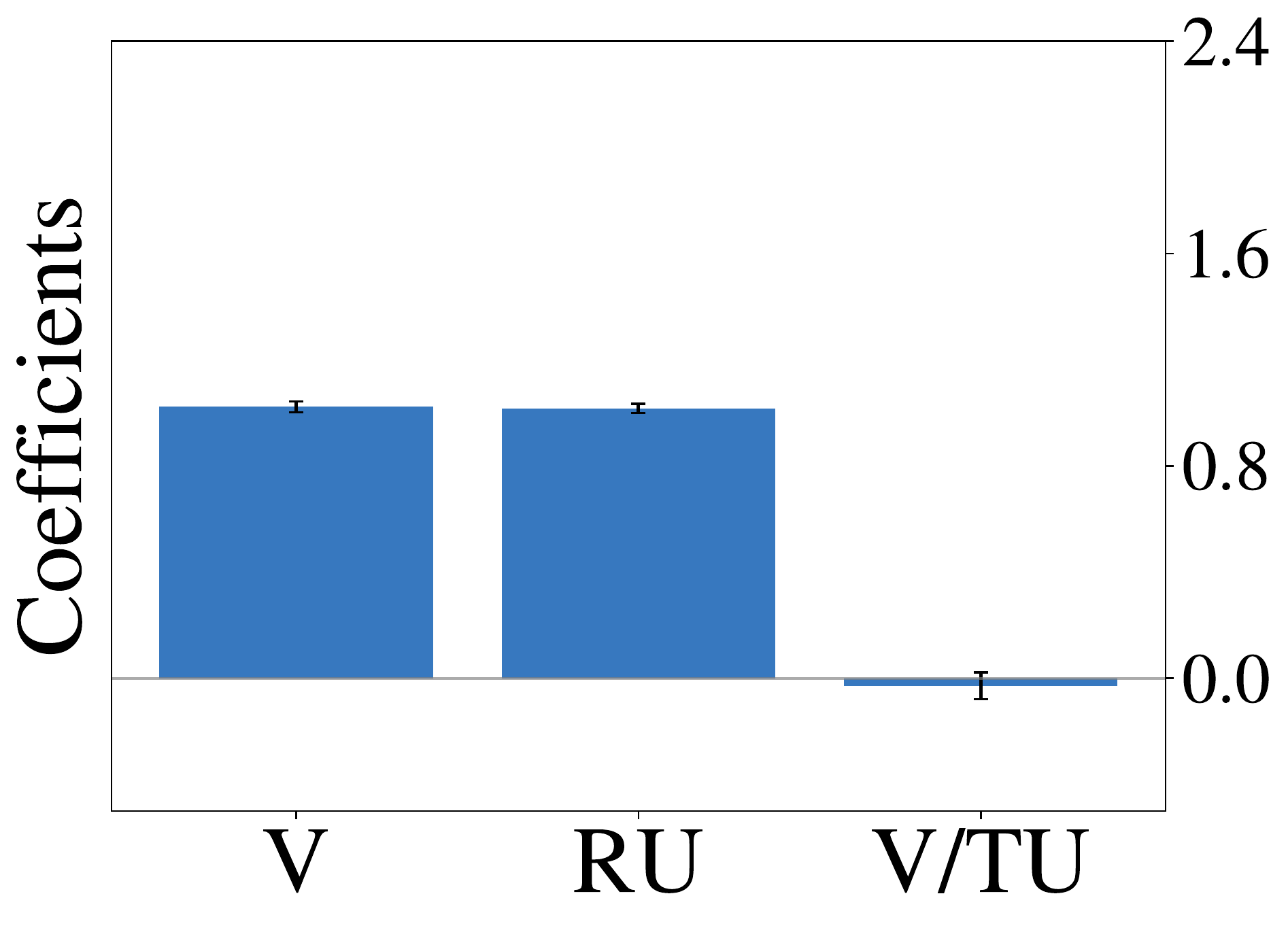} & \includegraphics[align=c, width=0.23\textwidth]{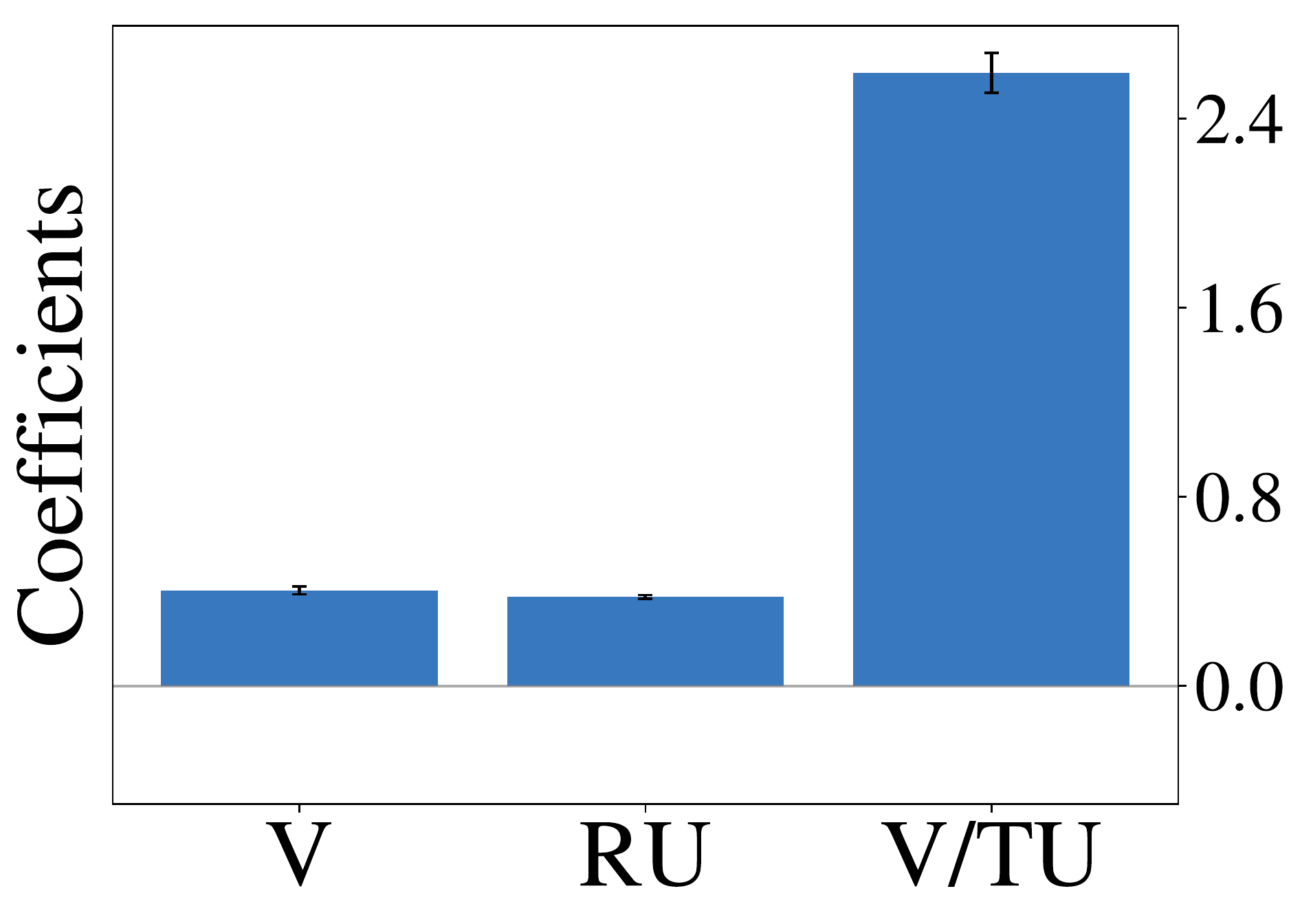}
    \end{tabular}
    
    \caption{Illustration of different algorithms for two-armed bandits. \textbf{Middle}: Definitions of the respective policy. \textbf{Bottom}: Coefficients obtained from fitting the probit regression (Equation \ref{eq:probit}) to corresponding trajectories. Error bars indicate the uncertainty (one standard deviation) in the coefficients estimated through a Laplace approximation. Note, that for LRLAs the coefficients are task-dependent. For this plot we use the set of two-armed bandits described in the later sections to compute the coefficients. $\Phi$ denotes the cumulative distribution function of a standard normal distribution.}
    \label{fig:baselines}
\end{figure*}

We maximize Equation \ref{eq:constraint} using standard gradient-based optimization techniques. For this we simulate $k$ environments in parallel and update the model at the end of each episode. All models in this work employ a group horseshoe prior, which can be viewed as a continuous relaxation of a spike-and-slab prior \cite{mitchell1988bayesian}, over their weights:
\begin{align*}
    &s \sim \mathcal{C}^+(0, \tau_0); ~~~ \tilde{z}_i \sim \mathcal{C}^+(0, 1); \\ 
    &\tilde{\theta}_{ij} \sim \mathcal{N}(0, 1); ~~~ \theta_{ij} = \tilde{\theta}_{ij}\tilde{z}_is
\end{align*}

and we represent the approximate posterior $q_{\phi}(\theta)$ through a fully factorized distribution as proposed in \cite{louizos2017bayesian}. The hyperparameter of the horseshoe prior is fixed to $\tau_0 = 10^{-5}$. The horseshoe prior is a sparsity-inducing prior, which causes our models to implement simple functions in absence of any experience. During training we approximate the expectation of the log-likelihood term with a single sample from $q_{\phi}(\theta)$ and make use of the reparametrization trick \cite{kingma2013auto}. Resampling of weight matrices is done only at the beginning of an episode as proposed in \citeA{gal2016theoretically, fortunato2017bayesian}. Target values $y_t$ are computed using the maximum a posteriori estimate of a separate target network \cite{mnih2013playing, lipton2017bbq}. For additional details we refer the reader to the publicly available implementation\footnote{\url{https://github.com/marcelbinz/MDLDQN}}. 

\subsection{Multi-Armed Bandits}

Experiments in the following section involve a multi-armed bandit task. These are MDPs consisting of a single state. At each step $t$ an agent selects one out of multiple actions and is rewarded according to an unknown, stationary distribution based on its choice. This interaction is repeated $T$ times. \\

The trade-off between exploiting good options and exploring yet unknown ones is the central theme in multi-armed bandits (and in RL in general). Methods for resolving this exploration-exploitation dilemma can be categorized in two major groups: directed and random exploration strategies. Directed exploration attempts to gather information about uncertain, but learnable, parts of the environment, while random exploration injects stochasticity of some form into the policy. \citeA{gershman2018deconstructing} showed, that these two principles can be distinguished exactly under certain conditions. For this we consider a two-armed bandit task with normal distributions over both the mean of rewards for each arm and their reward noise at each time-step. Let $\mathcal{N}(r_a; \mu_{0, a}, \sigma_{0, a})$ be an independent normal prior over expected rewards for each action $a$ and $\mathcal{N}(r_a; \mu_{t, a}, \sigma_{t, a})$ be the posterior after $t$ interactions. Many popular strategies can be formulated using the parameters of these distributions. Define:
\begin{align} \label{eq:factors}
    V_t &= \mu_{t, 0} - \mu_{t, 1} \nonumber \\ 
    RU_t &= \sigma_{t, 0} - \sigma_{t, 1}  \\
    TU_t &= \sqrt{\sigma_{t, 0}^2 + \sigma_{t, 1}^2} \nonumber
\end{align}

$V_t$ constitutes the estimated difference in value, while $RU_t$ and $TU_t$ describe relative and total uncertainty respectively. Choice probability in Thompson sampling (an example for random exploration) is only a function of $V_t$ and $TU_t$, while it is a function of $V_t$ and $RU_t$ for the UCB algorithm (an example of directed exploration). Figure \ref{fig:baselines} (middle row) shows definitions of all strategies under consideration. For a given set of observed trajectories $\mathcal{D}$ one can fit a probit regression model to infer the importance of factors from Equation \ref{eq:factors}:
\begin{equation} \label{eq:probit}
    p(a_t = 0 |\mathcal{D}, \mathbf{w}) = \Phi(w_1 {V}_t + w_2 {RU}_t + w_3 {V}_t/{TU}_t)
\end{equation}

Analyzing the resulting coefficients $\mathbf{w}$ can reveal, which exploration strategy generated the observations, as shown in Figure \ref{fig:baselines} (bottom row). We utilize this form of analysis throughout the following sections.

\section{Empirical Analysis}

\subsection{Human Participants}

\begin{figure*}[t]
    \centering
    \begin{minipage}{.18\textwidth}
    \centering
    \textbf{Examples}
    \includegraphics[width=\textwidth]{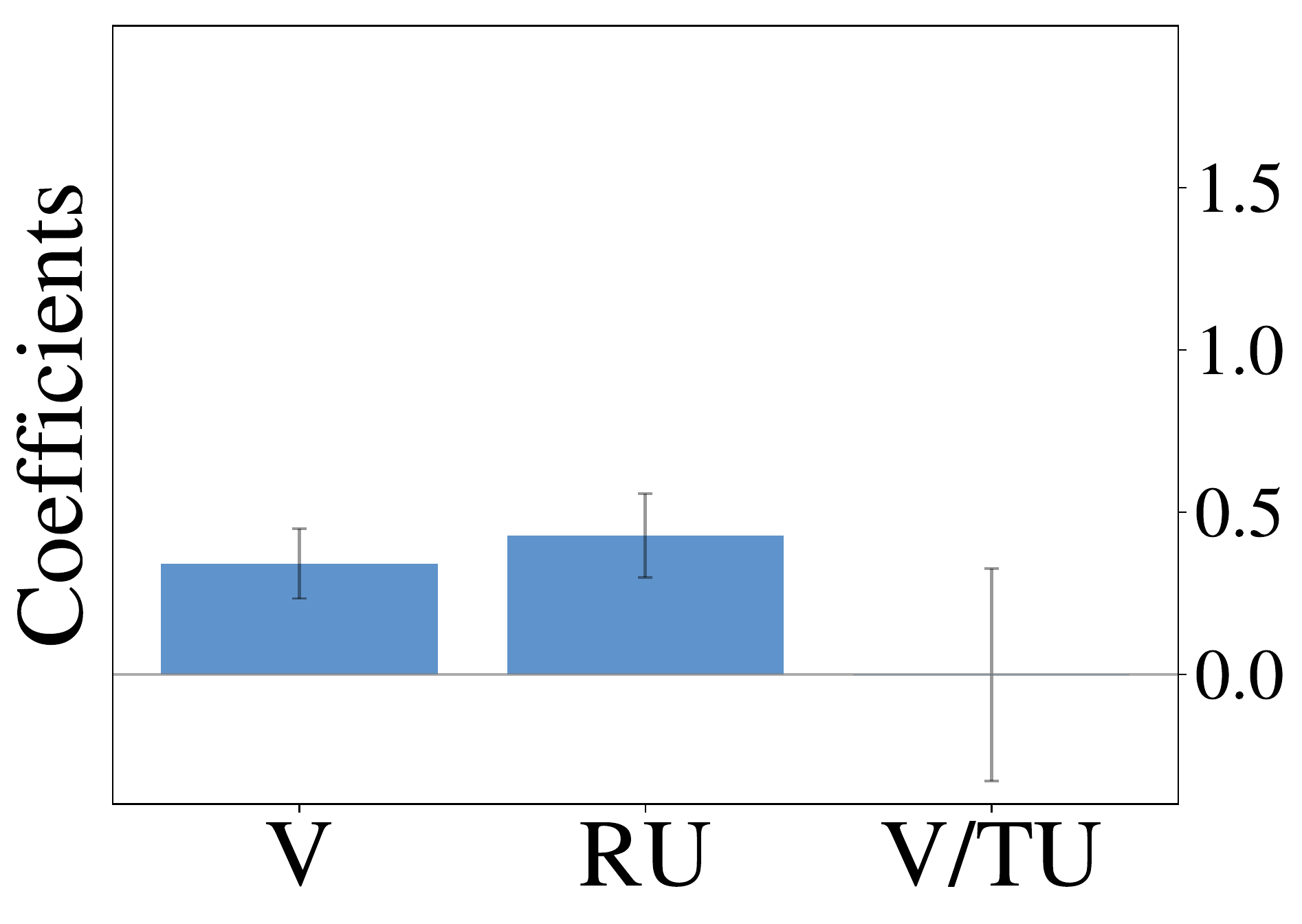}
    \includegraphics[width=\textwidth]{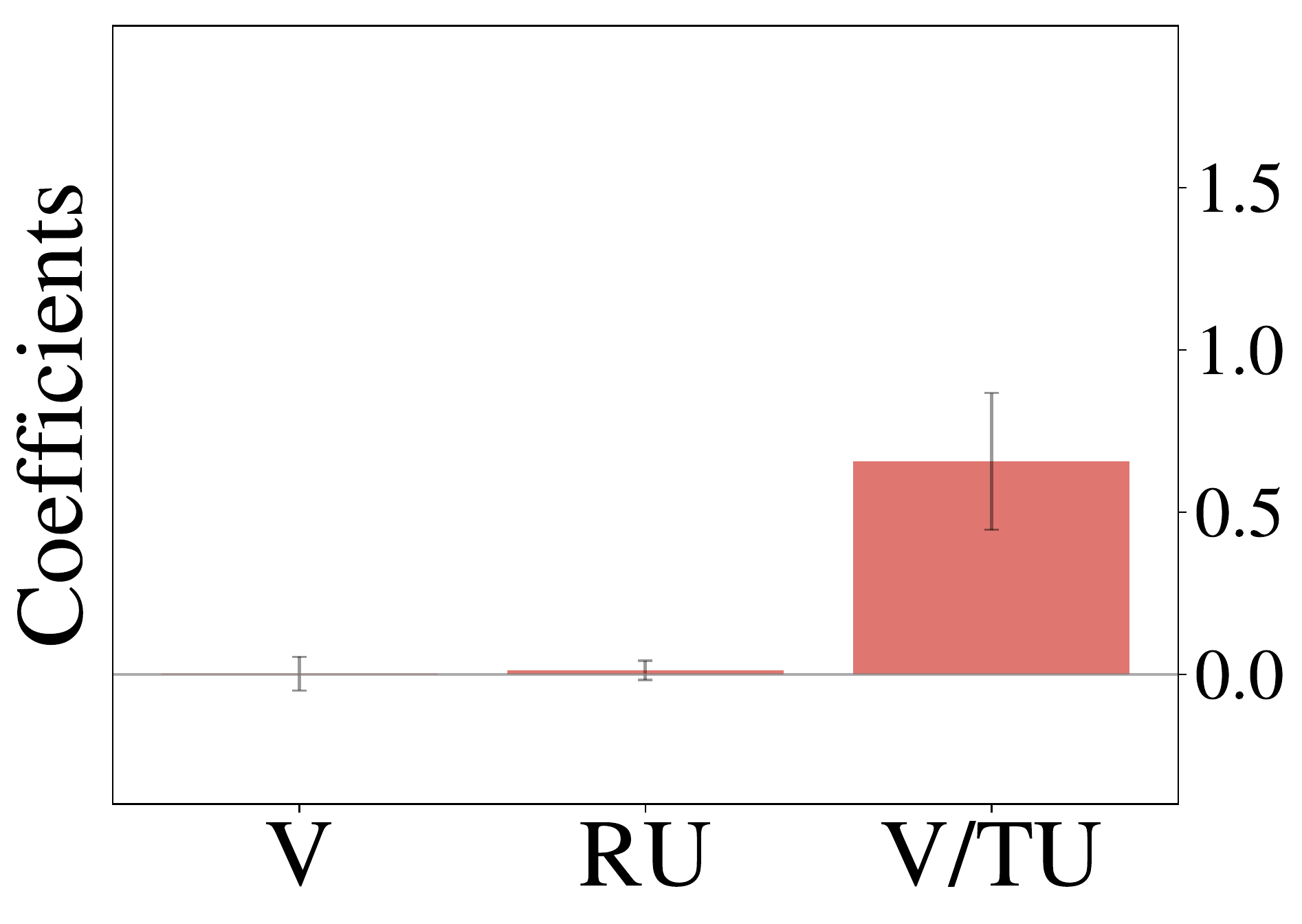}
    \includegraphics[width=\textwidth]{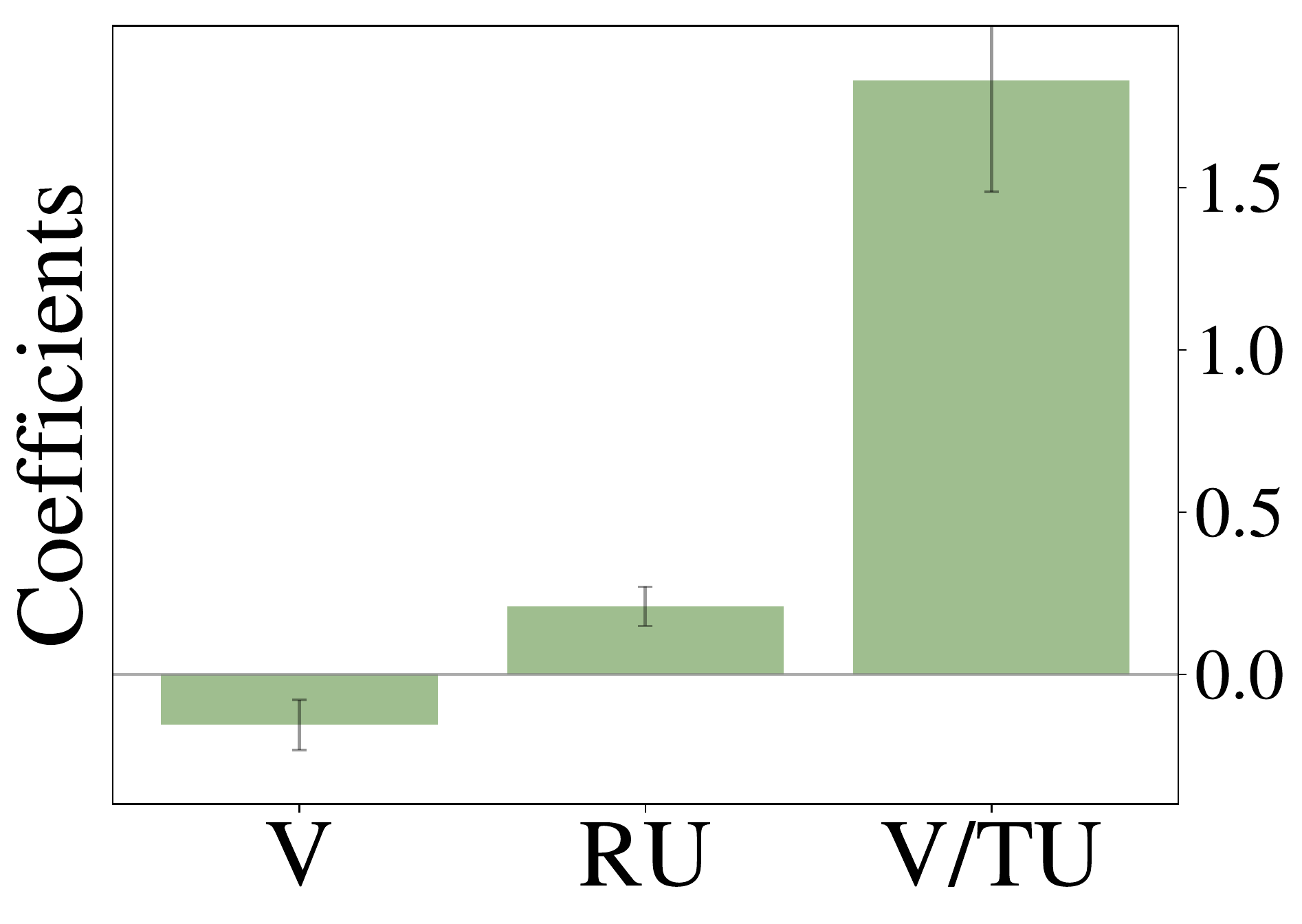}
    \end{minipage}
    \begin{minipage}{.4\textwidth}
    \centering
    \textbf{Humans}
    \includegraphics[width=\textwidth]{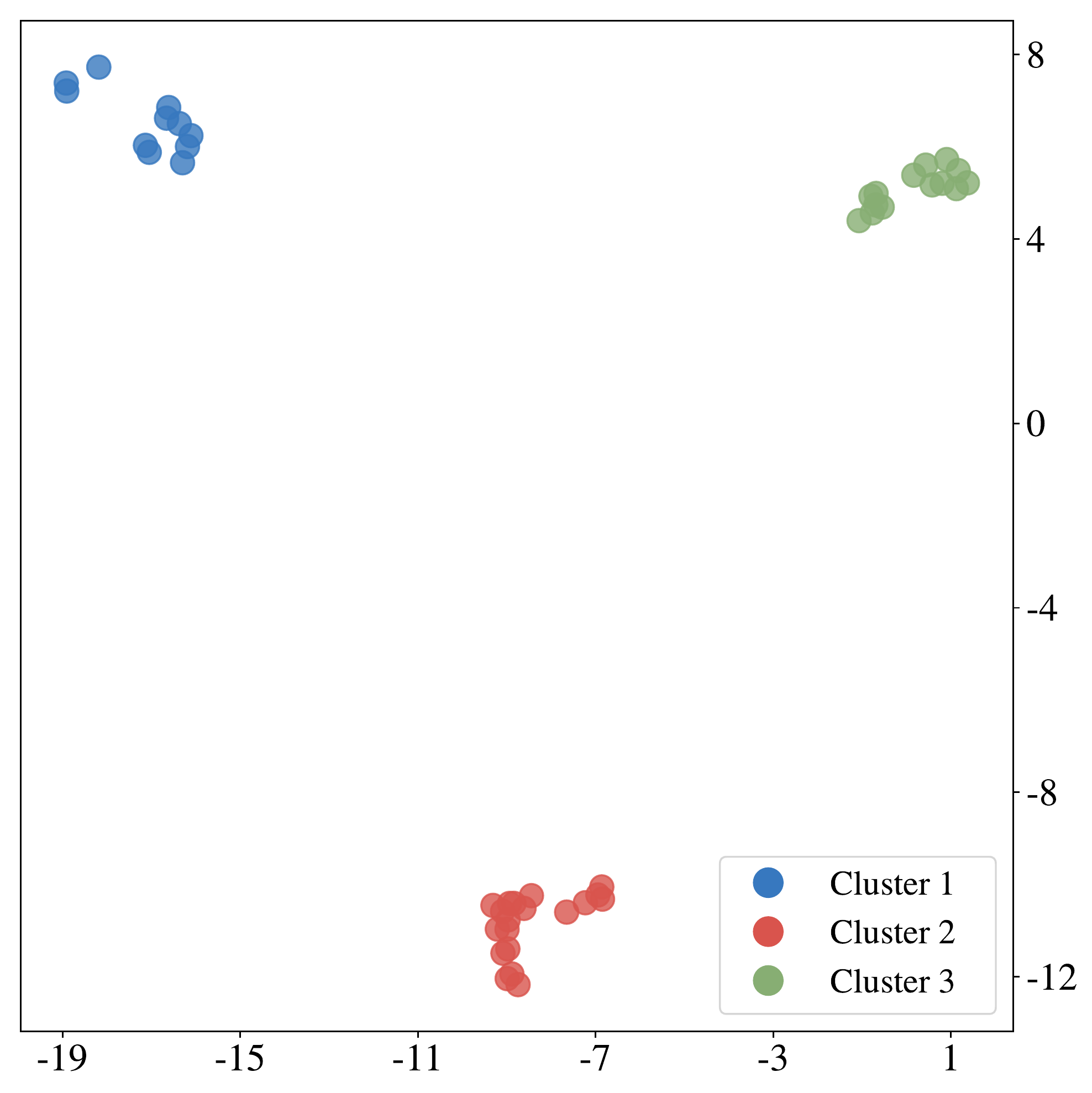}
    \end{minipage}
    \begin{minipage}{.4\textwidth}
    \centering
    \textbf{Human \& Models}
    \includegraphics[width=\textwidth]{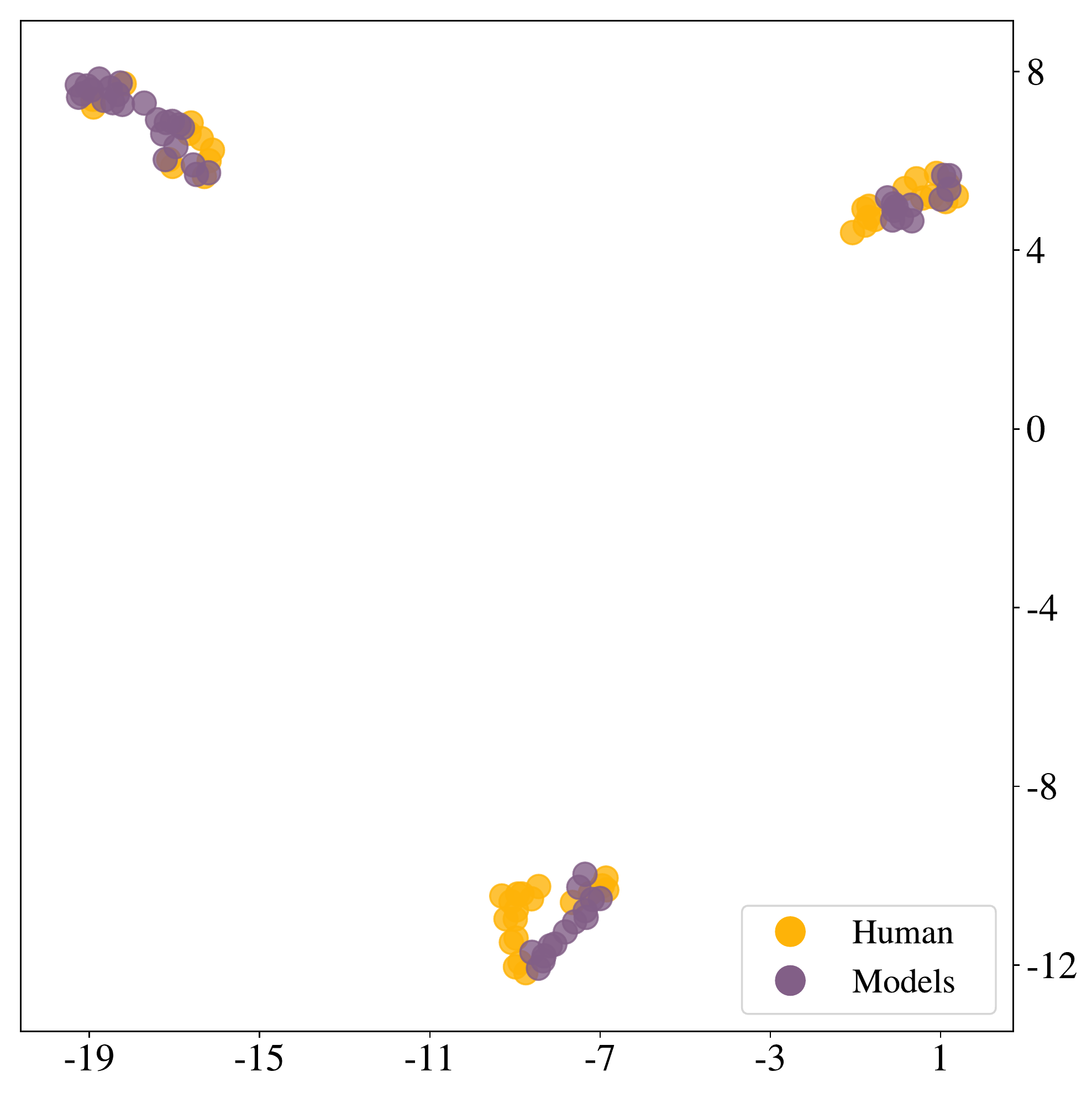}
    \end{minipage}
    
    \caption{Visualization of human policies alongside resource-constrained LRLAs. \textbf{Left}: Probit regression coefficients of prototype participants. Prototypes were obtained from a mean-shift clustering, shown in the middle plot. Colors correspond to clusters. Error bars indicate the uncertainty (one standard deviation) in the coefficients estimated through a Laplace approximation. \textbf{Middle}: UMAP \cite{2018arXiv180203426M} embedding of coefficients for all participants. \textbf{Right}: Joint UMAP embedding of coefficients for human participants and LRLAs $\in \mathcal{H}_{\text{LRLA}}$.}
    \label{fig:human_exp}
\end{figure*}

We initially inspect human exploration strategies on a two-armed bandit task with episode length $T = 10$. The mean reward for each action is drawn from $\mathcal{N}(\mu_a; 0, \sqrt{100})$ at the beginning of an episode and the reward in each step from $\mathcal{N}(r_t; \mu_{a_t}, \sqrt{10})$. Intuitively we expect some participants to be more proficient at the task, for example because they have more experience at related problems (higher $\hat{N}$), while the opposite is true for others. We rely on data gathered by \citeA{gershman2018deconstructing}, which contains records of 44 participants, each playing 20 of the aforementioned two-armed bandit problems. Figure \ref{fig:human_exp} (middle) shows the result of fitted probit regression coefficients for individual participants. This analysis reveals three major subgroups within the population, each using a different set of strategies. We visualize coefficients of three example participants (Figure \ref{fig:human_exp}, left) and observe, that a large fraction is well-described through Thompson sampling (clusters 2 and 3), while other participants have tendencies towards a mixture of strategies (cluster 1).

\subsection{Learned Reinforcement Learning Algorithms}

\begin{figure*}[t]
\centering
\begin{minipage}{.4\textwidth}
\centering
    \vspace*{0.5cm}
    \includegraphics[width=\textwidth]{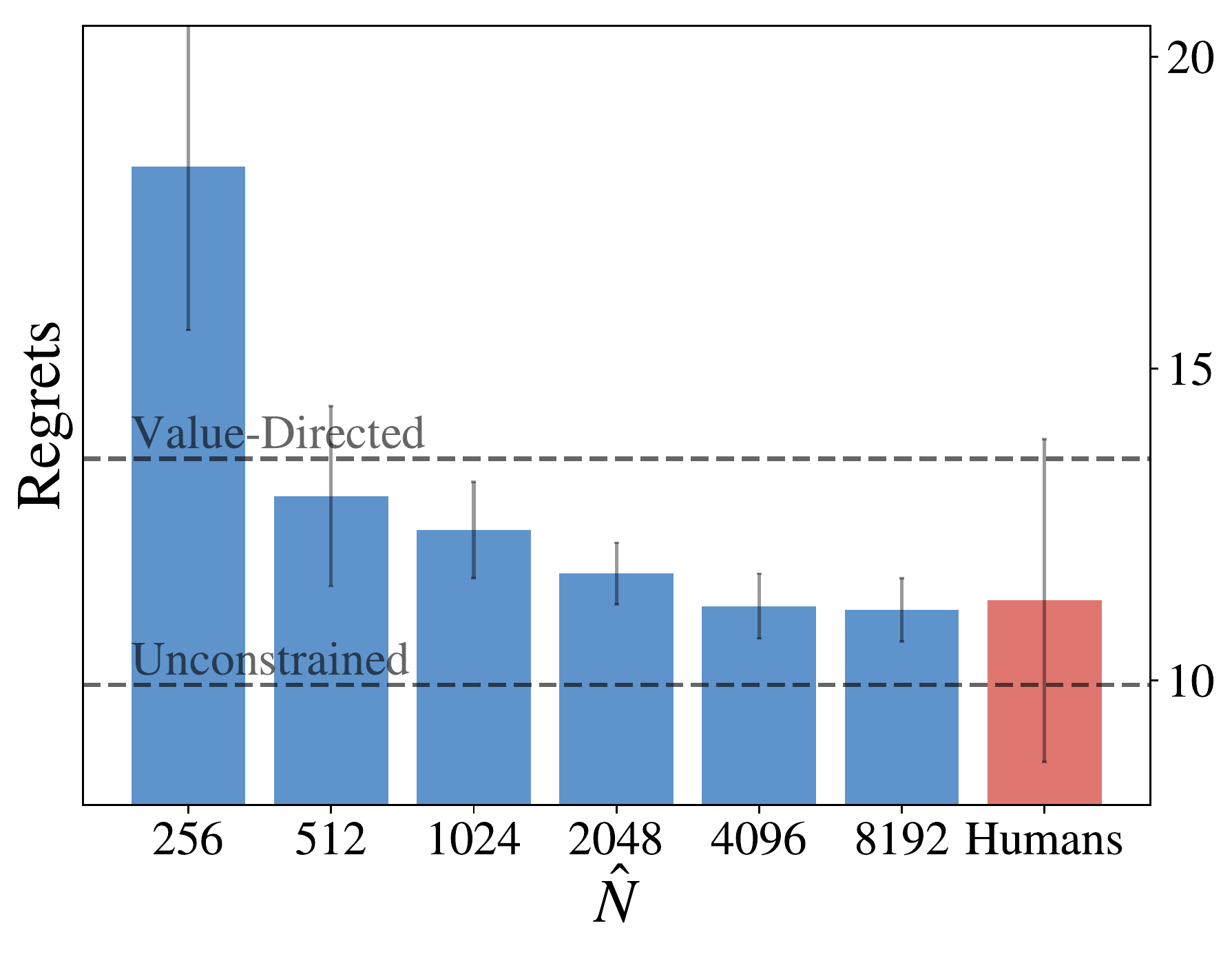}
\end{minipage}
\begin{minipage}{.18\textwidth}
\centering
\vspace*{-0.308cm}
    $\hat{N} = 256$
    \includegraphics[width=\textwidth]{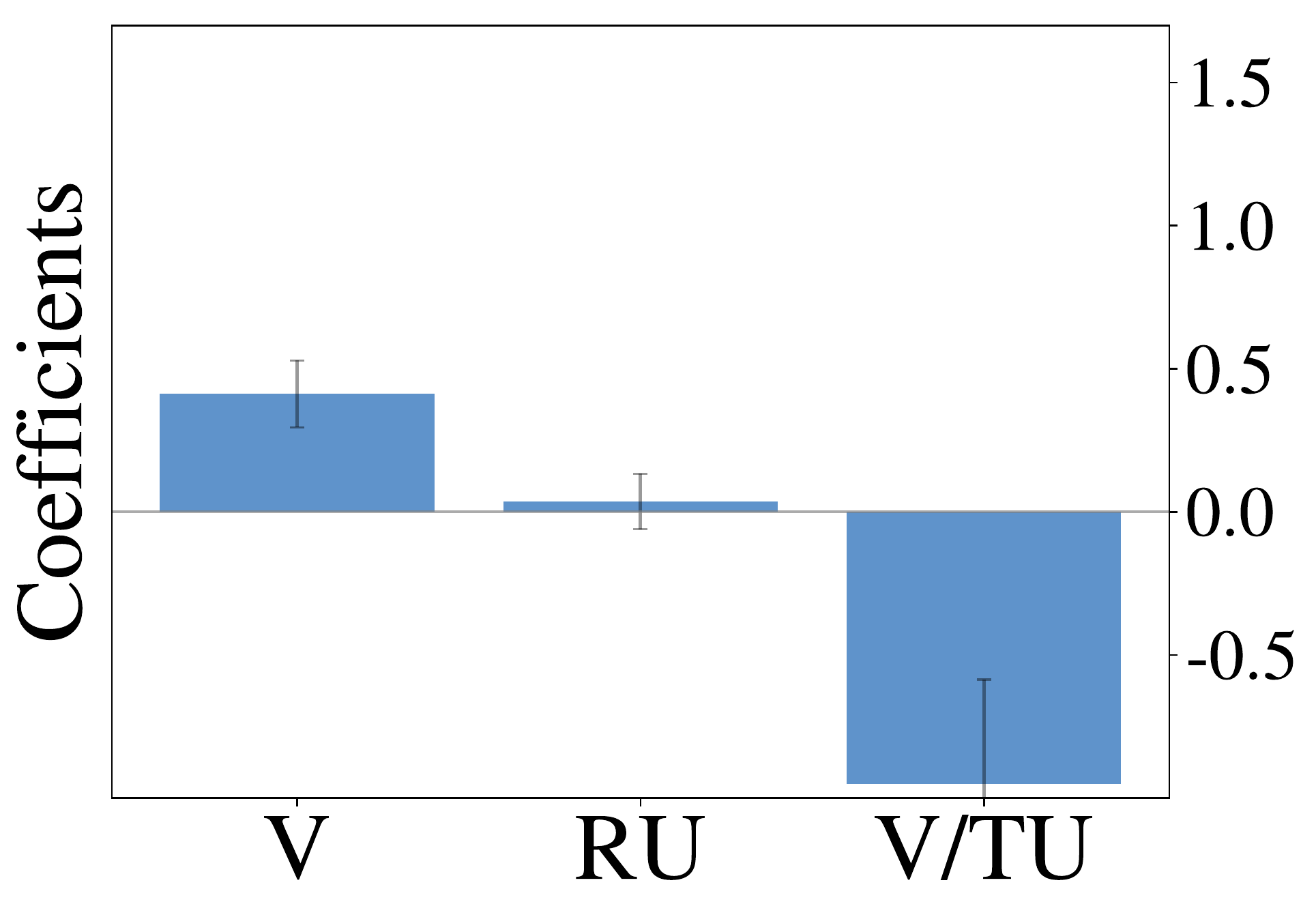}
    $\hat{N} = 2048$
    \includegraphics[width=\textwidth]{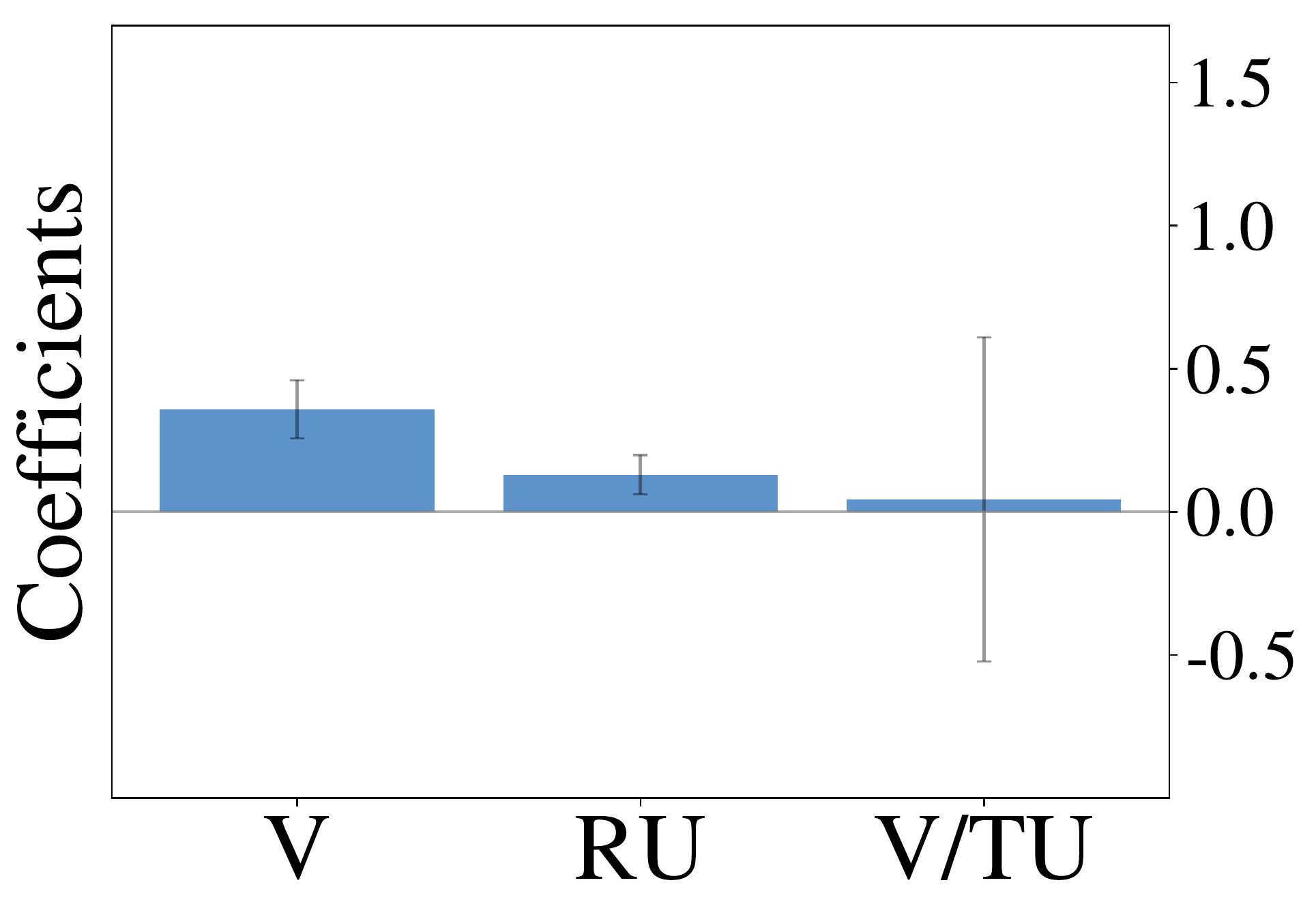}
\end{minipage}
\begin{minipage}{.18\textwidth}
\centering
\vspace*{-0.308cm}
    $\hat{N} = 512$
    \includegraphics[width=\textwidth]{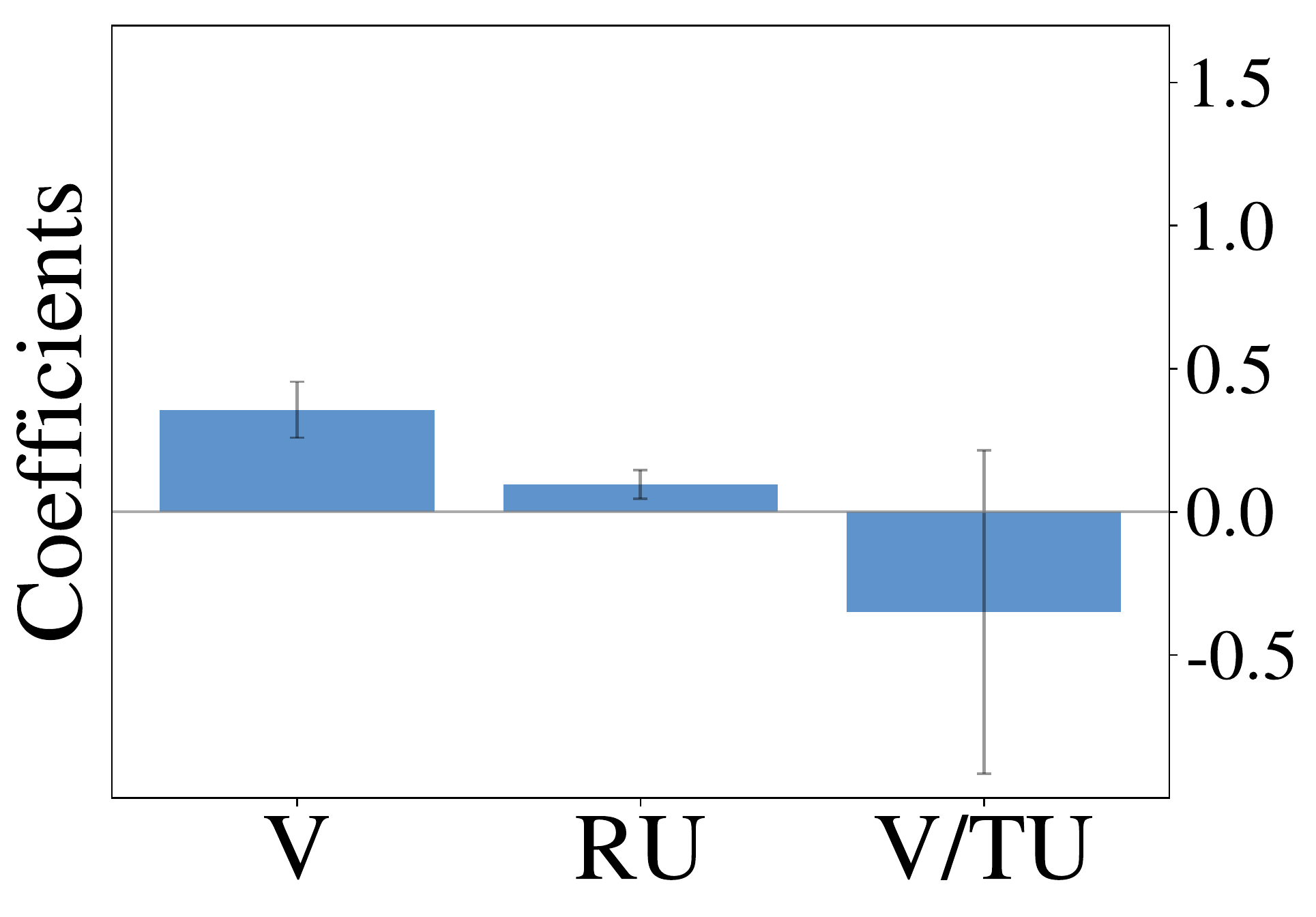}
    $\hat{N} = 4096$
    \includegraphics[width=\textwidth]{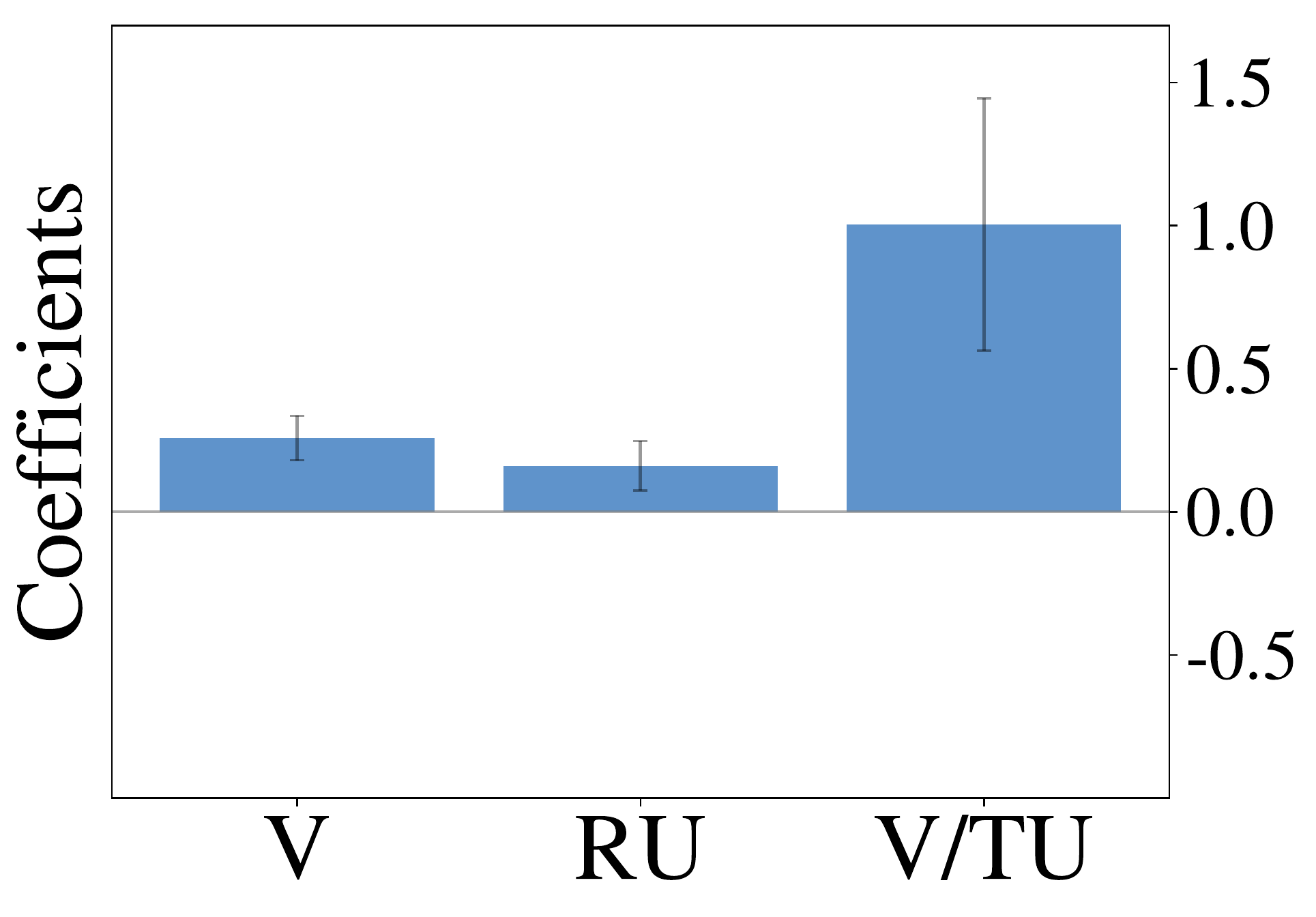}
\end{minipage}
\begin{minipage}{.18\textwidth}
\centering
\vspace*{-0.308cm}
    $\hat{N} = 1024$
    \includegraphics[width=\textwidth]{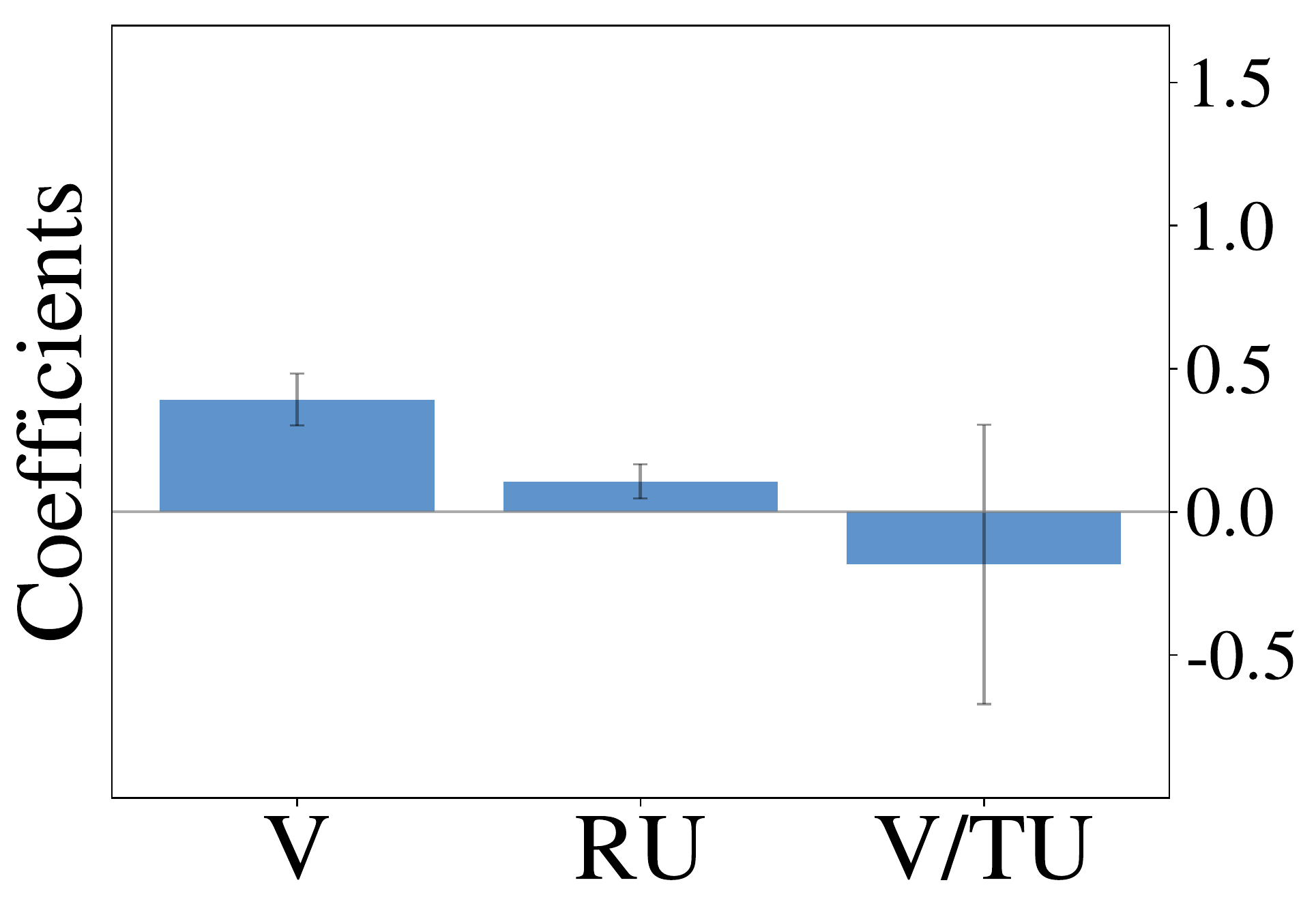}
    $\hat{N} = 8192$
    \includegraphics[width=\textwidth]{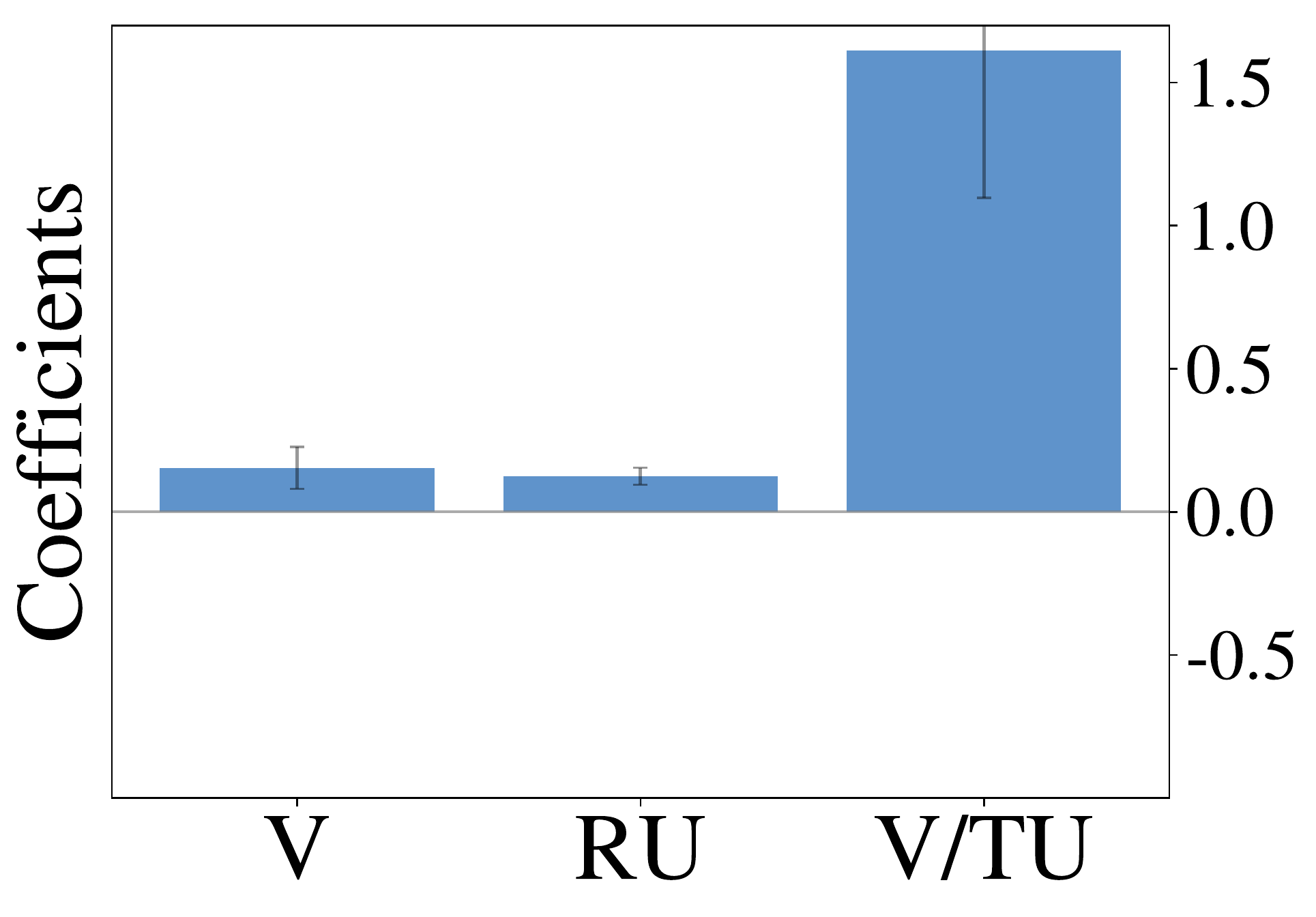}
\end{minipage}
\caption{Results for optimized LRLAs with different $\hat{N}$. \textbf{Left}: Visualization of per episode regret averaged over 10 models and 1000 episodes. Horizontal lines correspond to the performance of a value-directed policy and an unconstrained LRLA. \textbf{Right}: Coefficients of the probit regression from Equation \ref{eq:probit}. Error bars indicate standard deviations across the 10 models.}
    \label{fig:cont}
\end{figure*}

Next we show, that optimizing LRLAs with different regularization factors leads to the emergence of diverse exploration pattern. We train otherwise identical models for $\hat{N} \in \mathcal{H}_{\text{LRLA}} = \{256, 512, 1024, 2048, 4096, 8192 \}$ on the same two-armed bandit task until convergence and report average results over 10 random seeds unless otherwise noted. Equation \ref{eq:constraint} is approximated with a batch of samples from complete episodes of 16 parallel simulations and gradient-based optimization is performed using Adam \cite{kingma2014adam}. Figure \ref{fig:cont} (left) shows, that performances continuously improves as $\hat{N}$ increases, confirming our expectation that models become more sophisticated for large $\hat{N}$. Fitting the aforementioned probit regression model to the resulting policies (Figure \ref{fig:cont}, right) reveals value-based characteristics at one end of the spectrum. Towards the other end we observe coefficients, that slowly transition to those of the unconstrained ($\beta$ = 0) model.

\subsection{Modelling Human Behavior}

We are mainly interested in whether the set of resource-constrained LRLAs can help us to understand human behavior on an individual level. To answer this question, we compare the optimized models to human decision-making strategies in terms of the probit regression analysis. We visualize the regression coefficients for 50 models (10 for each value of $\hat{N} \in \mathcal{H}_{\text{LRLA}}$, excluding $\hat{N} = 256$) alongside those of the human participants in Figure \ref{fig:human_exp} (right). Although some parts of the low-dimensional embedding are over- and underrepresented, the overall variation of human exploration strategies is captured by the resource-constrained LRLAs.

\subsection{Model Comparison}

\begin{figure*}
    \centering
    
    \begin{minipage}{.55\textwidth}
    \centering
    \hspace*{1cm}\textbf{Log-likelihoods} \\
    \includegraphics[width=\textwidth]{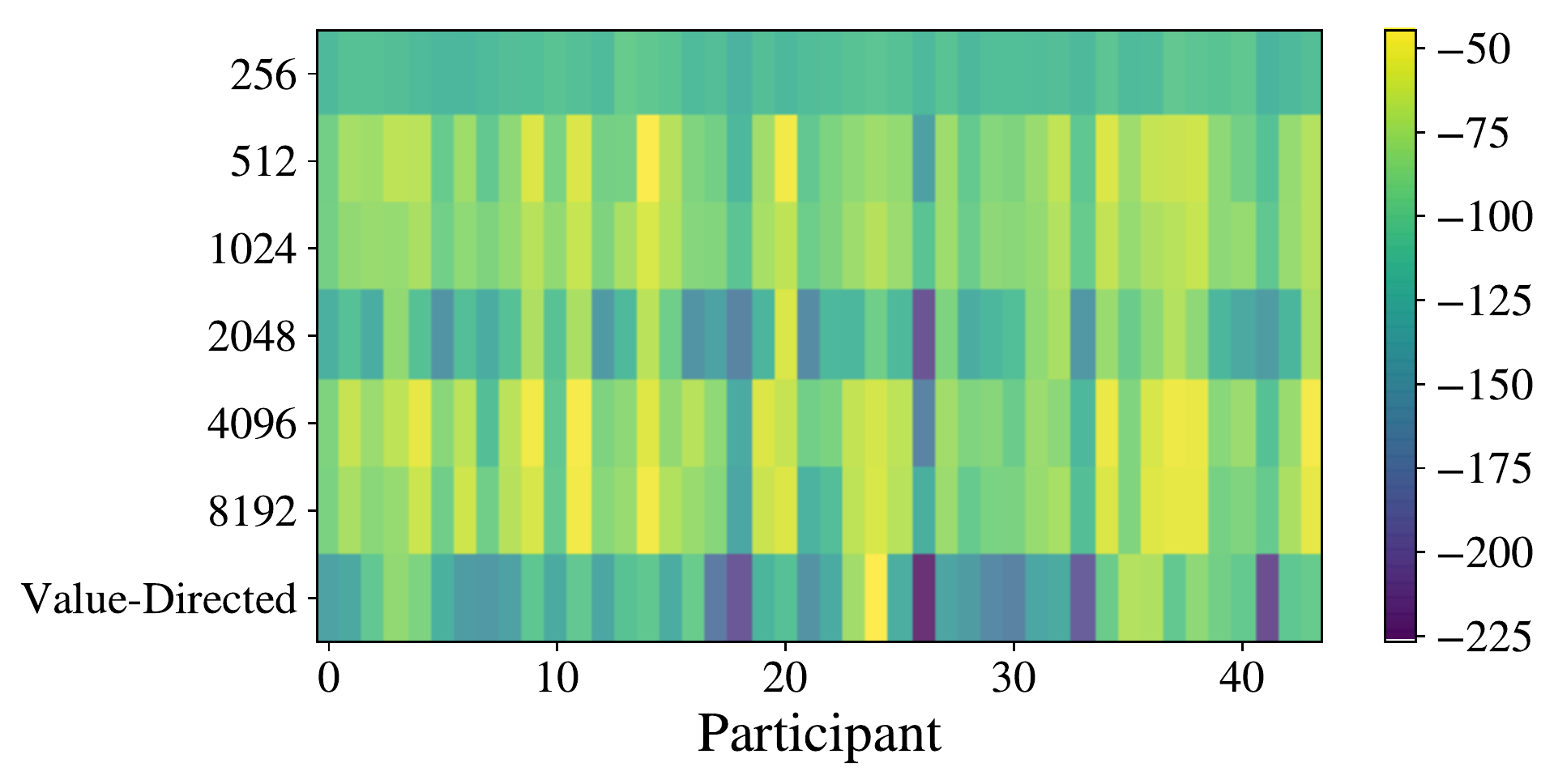}
    \end{minipage}
    \begin{minipage}{.42\textwidth}
    \centering
    \textbf{Bayes Factors} \\
    \includegraphics[width=\textwidth]{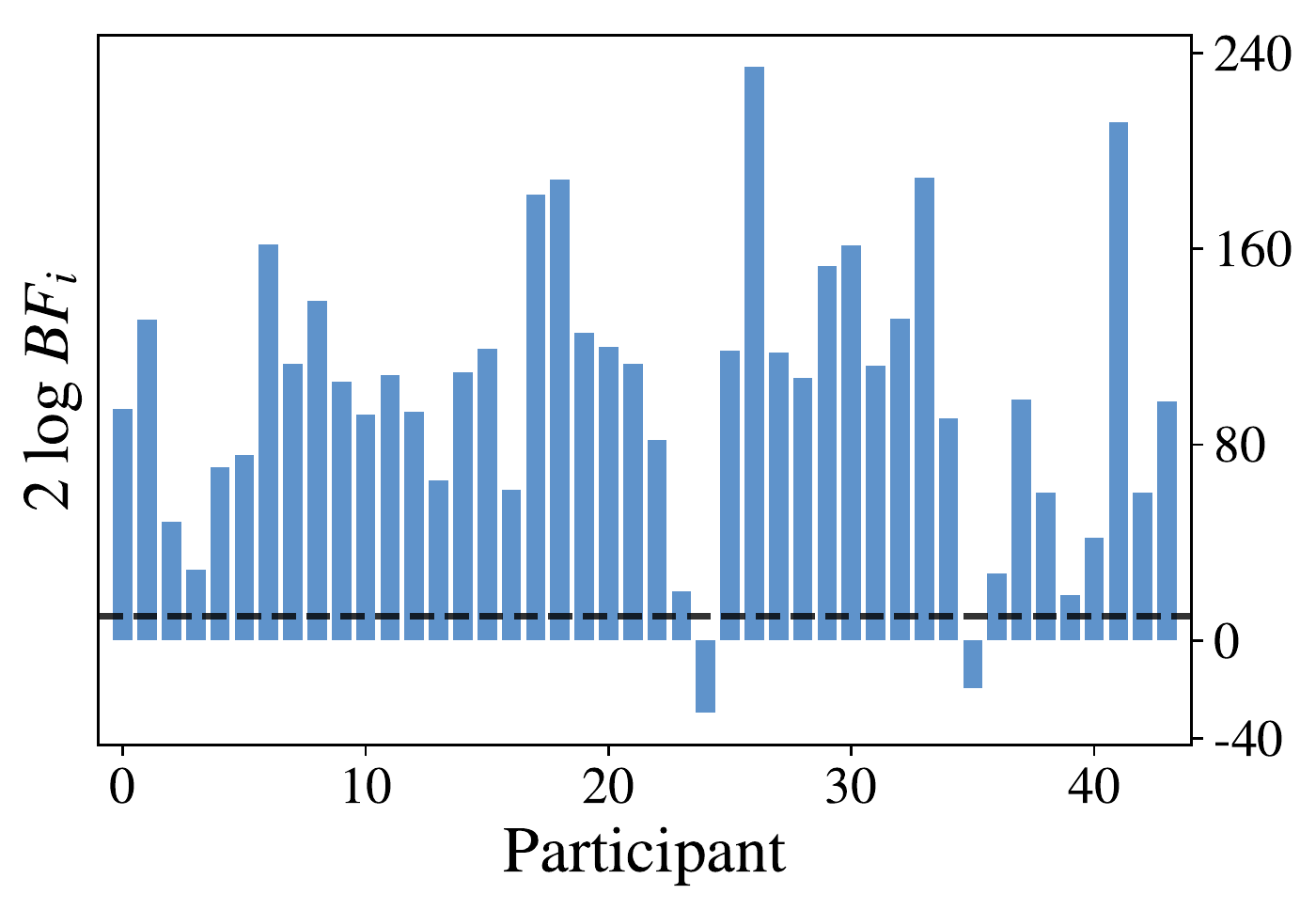}
    \end{minipage}
    \caption{Model comparison of the set of resource-constrained LRLAs with a value-directed baseline. \textbf{Left}: Log-likelihoods for each participant and model. Higher values indicate a better fit. \textbf{Right}: Bayes factors (see Equation \ref{eq:bf}) for each participant $i$. The dotted horizontal line (equal to 10) corresponds to the threshold for very strong evidence \cite{kass1995bayes} in favour of $\mathcal{H}_{\text{LRLA}}$.}
    \label{fig:comp}
\end{figure*}

The regression analysis performed so far provides only qualitative indicators for our hypothesis. In order to obtain a quantitative measure for the explanatory power of the proposed hypothesis, we performed a Bayesian model comparison. Figure \ref{fig:comp} (left) shows log-likelihoods for each participant and model. We observe, that different participants are modelled best with different values of $\hat{N}$. \\

To verify that the class of resource-constrained LRLAs $\mathcal{H}_{\text{LRLA}}$ contains a good model, we compute Bayes factors (BF) between  the marginal probability of the resource-constrained LRLAs and a value-directed policy:
\begin{align} \label{eq:bf}
    \log BF_i &= \log p(\mathcal{D}_i | \mathcal{H}_{\text{LRLA}}) - \log p(\mathcal{D}_i | H_{\text{value-directed}})  \nonumber \\
    p(\mathcal{D}_i | \mathcal{H}_{\text{LRLA}}) &= \frac{1}{|\mathcal{H}_{\text{LRLA}}|} \sum_{H \in \mathcal{H}_{\text{LRLA}}} p(\mathcal{D}_i | H)
\end{align}

where $\mathcal{D}_i$ refers to all actions taken by a specific participant and $\frac{1}{|\mathcal{H}_{\text{LRLA}}|}$ is a prior that corrects for multiple comparisons across different values of $\hat{N}$. The resulting $\log BF$s (see Figure \ref{fig:comp}, right) reveal strong evidence for 42 of the 44 participants in favor of the class of resource-constrained LRLAs, when compared with the baseline. This indicates, that one of the models in $\mathcal{H}_{\text{LRLA}}$ explains the participant's behavior much better than the value-directed policy. There are nine participants best described by letting $\hat{N} =512$, nine by $\hat{N} = 1024$, 20 by $\hat{N} = 4096$ and six by $\hat{N} = 8192$. This heterogeneity highlights, that the model class is able to accomodate individual differences between human participants. \\

Finally we want to show, that the proposed class of models captures exploration strategies across all participants better than any standard exploration strategy alone. To verify this, we computed Bayes factors between $\prod_i p(\mathcal{D}_i | \mathcal{H}_{\text{LRLA}})$ and two baseline exploration strategies: $\prod_i p(\mathcal{D}_i|H_{ \text{Thompson}})$ and $\prod_i p(\mathcal{D}_i|H_{ \text{UCB}})$. We find $2 \log BF = 72.8$ against Thompson sampling and $5391.4$ against UCB, indicating that our class of models is overall better at representing exploration strategies for all participants in comparison to any single, fixed strategy.

\section{Discussion}

In this work we proposed a justification for seemingly sub-optimal human strategies in sequential decision-making problems based on the idea of computational rationality. We view human decision-making as an instance of a learned, resource-constrained RL algorithm. This is formalized through learning distributions over parameters of a meta-learning model with a regularized, resource-rational objective. The emerging spectrum of strategies resembles characteristics of human decision-making without being explicitly trained to do so. Additional model comparison suggests, that the resulting resource-constrained LRLAs describe human policies well on a quantitative level. However, the correspondence between human behavior and the LRLA model class is not perfect. Looking at Figure \ref{fig:human_exp} (right) we observe, that some clusters are not represented exactly. Furthermore it remains open, why none of the participants is best described through the model with $\hat{N} = 2048$.  Accounting for these observations remains a question for future work.  \\

The analysis on the two-armed bandit task presented in this work can be extended in several ways. Relating deliberation times to regularization factors could, for example, provide additional evidence for our hypothesis. It also remains to be seen whether our conclusions transfer to other sequential decision-making problems beyond the bandit setting. In this context we are especially interested in tasks, where descriptive models of individual human behavior consist of a set of different heuristics. We are also interested in methods, that allow us to disentangle resource-rational behavior from the Bayesian interpretation. \\

Recent work on model-free meta-learning methods, similar to the one employed in this work, indicates an emergence of model-based behavior \cite{wang2016learning} and causal reasoning \cite{dasgupta2019causal} as well as the ability for few-shot learning \cite{santoro2016meta}, properties supposedly absent in artificial systems. Having systems capable of such feats, opens the possibility for interesting studies on human cognition.

\vfill 

\section{Acknowledgments}

This work was supported by the DFG GRK-RTG 2271 'Breaking Expectations'.

\bibliographystyle{apacite}

\setlength{\bibleftmargin}{.125in}
\setlength{\bibindent}{-\bibleftmargin}

\bibliography{cogsci_template}

\end{document}